\documentclass[10pt,twocolumn,letterpaper]{article}

\usepackage{iccv}
\usepackage{times}
\usepackage{epsfig}
\usepackage{graphicx}
\usepackage{amsmath}
\usepackage{amssymb}
\usepackage{subfigure}
\usepackage{multirow}
\usepackage{tabularx}
\usepackage{color}
\usepackage{lipsum}

\newcommand\blfootnote[1]{%
  \begingroup
  \renewcommand\thefootnote{}\footnote{#1}%
  \addtocounter{footnote}{-1}%
  \endgroup
}

\iccvfinalcopy 

\def\iccvPaperID{1894} 

\ificcvfinal\fi
\begin{document}
\title{Selecting Relevant Web Trained Concepts for Automated Event Retrieval}

\author{Bharat Singh*, Xintong Han*, Zhe Wu, Vlad I. Morariu and Larry S. Davis\\
Center For Automation Research, University of Maryland, College Park\\
{\tt\small {\{bharat,xintong,zhewu,morariu,lsd\}}@umiacs.umd.edu}
}

\maketitle
\blfootnote{*The first two authors contributed equally to this paper.}
\thispagestyle{empty}

\begin{abstract}
Complex event retrieval is a challenging research problem, especially when no training videos are available. An alternative to collecting training videos is to train a large semantic concept bank a priori. Given a text description of an event, event retrieval is performed by selecting concepts linguistically related to the event description and fusing the concept responses on unseen videos. However, defining an exhaustive concept lexicon and pre-training it requires vast computational resources. Therefore, recent approaches automate concept discovery and training by leveraging large amounts of weakly annotated web data. Compact visually salient concepts are automatically obtained by the use of concept pairs or, more generally, n-grams. However, not all visually salient n-grams are necessarily useful for an event query--some combinations of concepts may be visually compact but irrelevant--and this drastically affects performance. We propose an event retrieval algorithm that constructs pairs of automatically discovered concepts and then prunes those concepts that are unlikely to be helpful for retrieval. Pruning depends both on the query and on the specific video instance being evaluated. Our approach also addresses calibration and domain adaptation issues that arise when applying concept detectors to unseen videos. We demonstrate large improvements over other vision based systems on the TRECVID MED 13 dataset.
\end{abstract}

\section{Introduction}
 Complex event retrieval from databases of videos is difficult because in addition to the challenges in modeling the appearance of static visual concepts--e.g., objects, scenes--modeling events also involves modeling temporal variations. In addition to the challenges of representing motion features and time, one particularly pernicious challenge is that the number of potential events is much greater than the number of static visual concepts, amplifying the well-known long-tail problem associated with object categories. Identifying and collecting training data for a comprehensive set of objects is difficult. For complex events, however, the task of even enumerating a comprehensive set of events is daunting, and collecting curated training video datasets for them is entirely impractical.

Consequently, a recent trend in the event retrieval community is to define a set of simpler visual concepts that are practical to model and then combine these concepts to define and detect complex events. This is often done when no examples of the complex event of interest are available for training. In this setting, training data is still required, but only for the more limited and simpler concepts. For example, \cite{chen2014event,liu2013video} discover and model concepts based on single words or short phrases, taking into account how {\em visual} the concept is. Others model pairs of words or n-grams in order to disambiguate between the multiple visual meanings of a single word \cite{divvalalearning} and take advantage of co-occurrences present in the visual world \cite{mensinkcosta}. An important aspect of recent work \cite{wu2014zero,chen2014event} is that concept discovery and training set annotation is performed automatically using weakly annotated web data. Event retrieval is performed by selecting concepts linguistically related to the event description and computing an average of the concept responses as a measure for event detection.
\begin{figure*}[htbp]
\begin{center}
\includegraphics[width=1\textwidth]{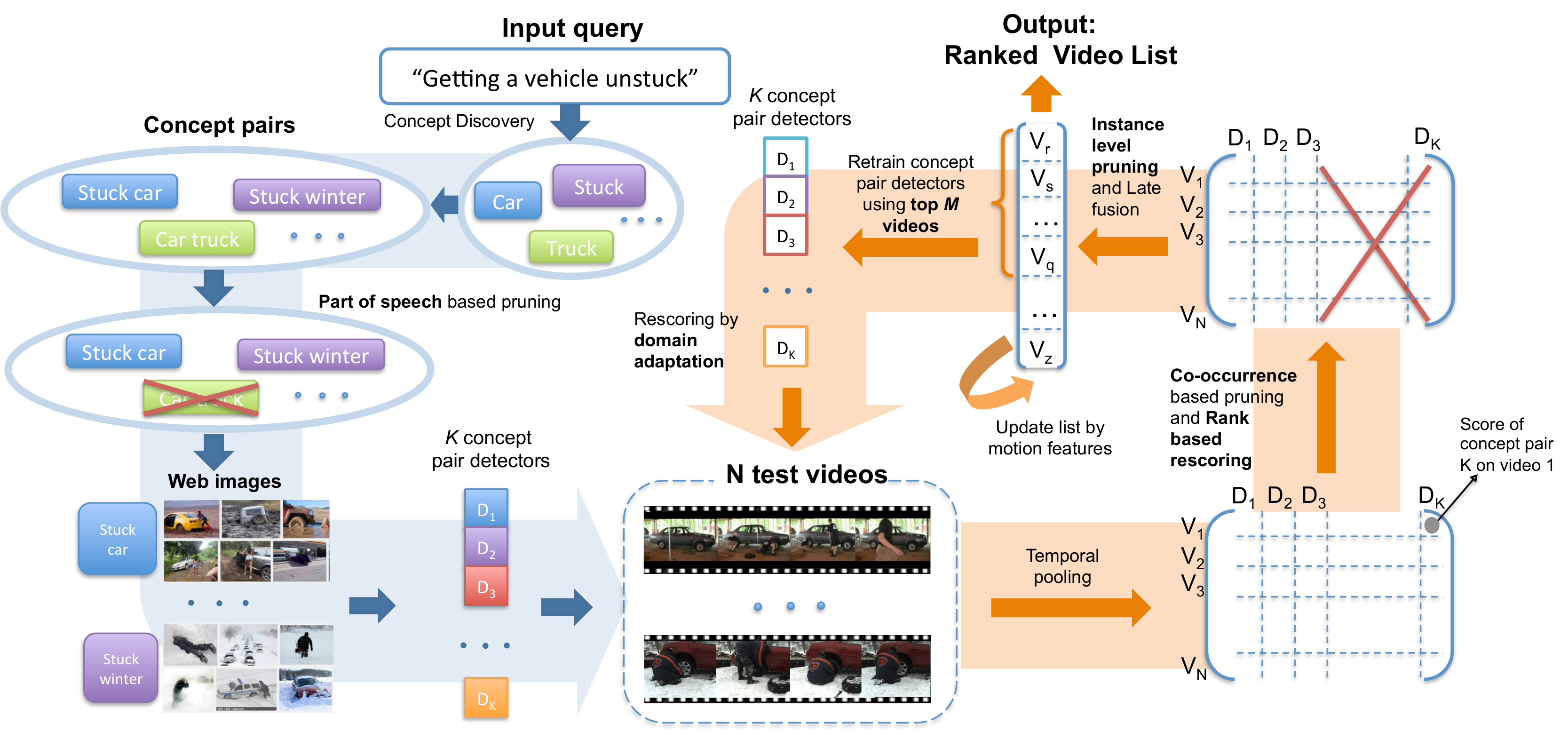}
\caption{Framework overview. An initial set of concepts is discovered from the web and transformed to concept pairs using an action centric part of speech (grammar) model. These concept pairs are used as Google Image search text queries, and detectors are trained on the search results. Based on the detector scores on the test videos, co-occurrence based pruning removes concepts that are likely to be outliers. Detectors are calibrated using a rank based re-scoring method. An instance level pruning method determines how many concepts are likely to be observed in a video and discards the lowest scoring concepts. The scores of remaining concepts are fused to score each video. Motion features of the top ranked videos are used to train a SVM and update the video list. Finally, the initial detectors are re-trained using the top ranked videos of this video list, and the process of co-occurrence based pruning, instance level pruning and rank based calibration is repeated to re-score the videos.}
\vspace{-10px}
\end{center}
\label{fig:framework}
\end{figure*}

Based on recent advances, we describe a system that ranks videos based on their similarity to a textual description of a complex event, using only web resources and without additional human supervision. In our approach, the textual description is represented by and detected through a set of concepts. Our approach builds on \cite{chen2014event} for discovering concepts given a textual description of a complex event, and \cite{divvalalearning} for automatically replacing the initial concepts with {\em concept pairs} that are visually salient and capture specific visual meanings. 

However, we observe that many visually salient concepts generated from an event description are not useful for detecting the event. In fact, we find that removing certain concepts is a key step that significantly improves event retrieval performance. Some concepts should be removed at training time because they model visually salient concepts that are not likely to be meaningful based on linguistic considerations. Others should be removed if an analysis of video co-occurrences and activation patterns indicates that a concept is likely to be irrelevant or not among the subset of concepts that occur in a video instance. These problems are further confounded by the fact that concept detectors are initially trained on weakly supervised web images\footnote{We prefer to use web images for concept training because a web search is a weak form of supervision which provides no spatial or temporal localization. This means that if we search for video examples of a concept, we do not know how many and which frames contain the concept (a temporal localization issue), while an image result is much more likely to contain the concept of interest (the spatial localization still remains).}, so there is a domain shift to video, and detector responses are not properly calibrated.

Our contribution is a fully automatic algorithm that discovers concepts that are not only visually salient, but are also likely to predict complex events by exploiting co-occurrence statistics and activation patterns of concepts. We address domain adaptation and calibration issues in addition to modelling the temporal properties. Evaluations are conducted using the TRECVID EK0 dataset, where our system outperforms state-of-the-art methods based on visual information.

\section{Related Work}


Large scale video retrieval commonly employs a concept-based video representation (CBRE) \cite{assari2014video,mazloom2013querying,merler2012semantic,yu2012multimedia}, especially when only few or no training examples of the events are available. In this setting, complex events are represented in terms of a large set of concepts that are either event-driven (generated once the event description is known) \cite{chen2014event,habibian2014composite,liu2013video} or pre-defined \cite{wu2014zero,cui2014building,dalton2013zero}. A test query description is mapped to a set of concepts whose detectors are then applied to videos to perform retrieval. However, methods based on pre-defined concepts need to train an exhaustive set of concept detectors a priori or the semantic gap between the query description and the concept database might be too large. This is computationally expensive and currently infeasible for real-world video retrieval systems. Instead, in this paper, given the textual description of the event to be retrieved, our approach leverages web image data to discover event-driven concepts and train detectors that are relevant to this specific event.

Recently, web (Internet) data has been widely used for knowledge discovery \cite{duan2012visual, berg2010automatic, divvalalearning, wu2014zero, duan2012exploiting, habibian2014videostory}. Chen et al. \cite{chen2013neil} use web data to weakly label images, learn and exploit common sense relationships. Berg  et al. \cite{berg2010automatic} automatically discover attributes from unlabeled Internet images and their associated textual descriptions. Duan et al. \cite{duan2012visual} describe a system that uses a large amount of weakly labeled web videos for visual event recognition by measuring the distance between two videos and a new transfer learning method. Habibian et al. \cite{habibian2014videostory} obtain textual descriptions of videos from the web and learn a multimedia embedding for few-example event recognition. For concept training, given a list of concepts, each corresponding to a word or short phrase, web search is commonly used to construct weakly annotated training sets \cite{chen2014event, wu2014zero, divvalalearning}. We use the concept name as a query to a search engine, and train the concept detector based on the returned images.

Moreover, retrieval performance depends on high quality concept detectors. While the performance of a concept detector can be estimated (e.g., by cross-validation \cite{divvalalearning}), ambiguity remains in associating linguistic concepts to visual concepts. For example, {\em groom} in {\em grooming an animal} and {\em groom} in {\em wedding ceremony} are totally different, and while two separate detectors might be capable of modeling both types of {\em groom} separately, a single {\em groom} detector would likely perform poorly. Similarly, tire images from the web are different from frames containing tires in a video about {\em changing a vehicle tire}, since there are often people and cars in these frames. To solve this problem, \cite{divvalalearning,li2011composing} use an n-gram model to differentiate between multiple senses of a word. Habibian et al. \cite{habibian2014composite} instead leverage logical relationships (e.g., ``OR'', ``AND'', ``XOR'') between two concepts. Mensink et al. \cite{mensinkcosta} exploit label co-occurrence statistics to address zero-shot image classification. However, it is not sufficient to discover visually distinctive concepts, since not all concepts are equally informative for modeling events. We present a pruning process to discover visually distinctive {\em and useful} concepts by a pruning process.

Recent work has also explored multiple modalities--e.g., automatic speech recognition (ASR), optical character recognition (OCR), audio, and vision--for event detection \cite{jiang2014easy,jiang2014zero,wu2014zero} to achieve better performance over vision alone. Jiang et al. \cite{jiang2014zero} propose MultiModel Pseudo Relevance Feedback (MMPRF), which selects several feedback videos for each modality to train a joint model. Applied to test videos, the model yields a new ranked video list that is used as feedback to retrain the model. Wu et al. \cite{wu2014zero} represent a video by using a large concept bank, speech information, and video text. These features are projected to a high-dimensional concept space, where event/video similarity scores are computed to rank videos.  While multi-modal techniques achieve good performance, their visual components alone significantly under-perform the system as a whole.

All these methods suffer from calibration and domain adaptation issues, since CBRE methods fuse multiple concept detector responses and are usually trained and tested on different domains. To deal with calibration issues, most related work uses SVMs with probabilistic outputs \cite{lin2007note}. However, the domain shift between web training data and test videos is usually not addressed by calibration alone. To reduce this effect, some ranking-based re-scoring schemes \cite{jiang2014easy,jiang2014zero} replace raw detector confidences with the confidence rank in a list of videos. To further adapt to new domains (e.g., from images to videos), easy samples have been used to update detector models \cite{tang2012shifting,jiang2014easy}. Similar to these approaches, we use a rank based re-scoring scheme to address calibration issues and update models using the most confident detections to adapt to new domains.
\begin{figure*}[]
\begin{center}

    \label{fig:jump+bicycle}
   \includegraphics[height = 0.142\linewidth]{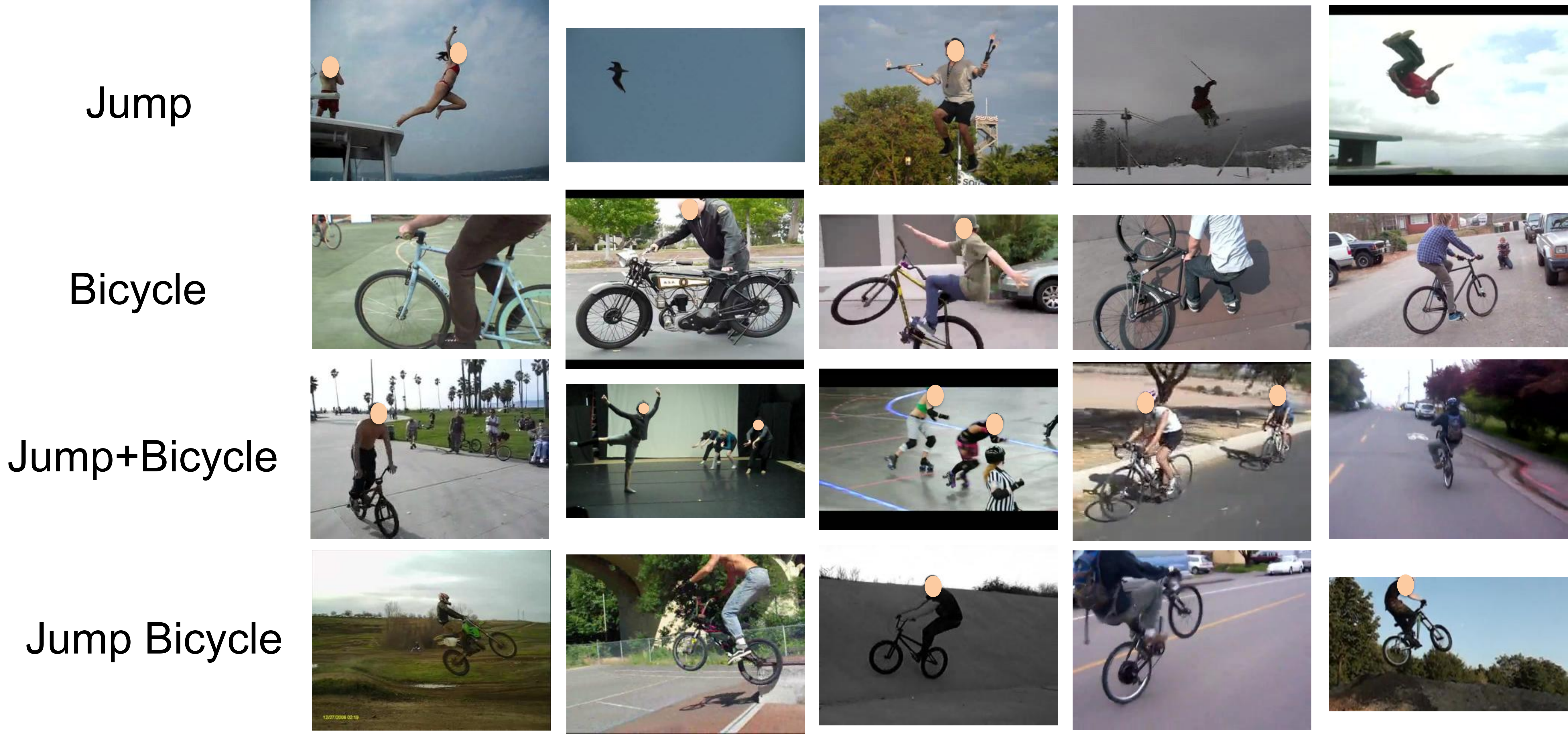}
    \label{fig:changing+tire}
   \includegraphics[height = 0.142\linewidth]{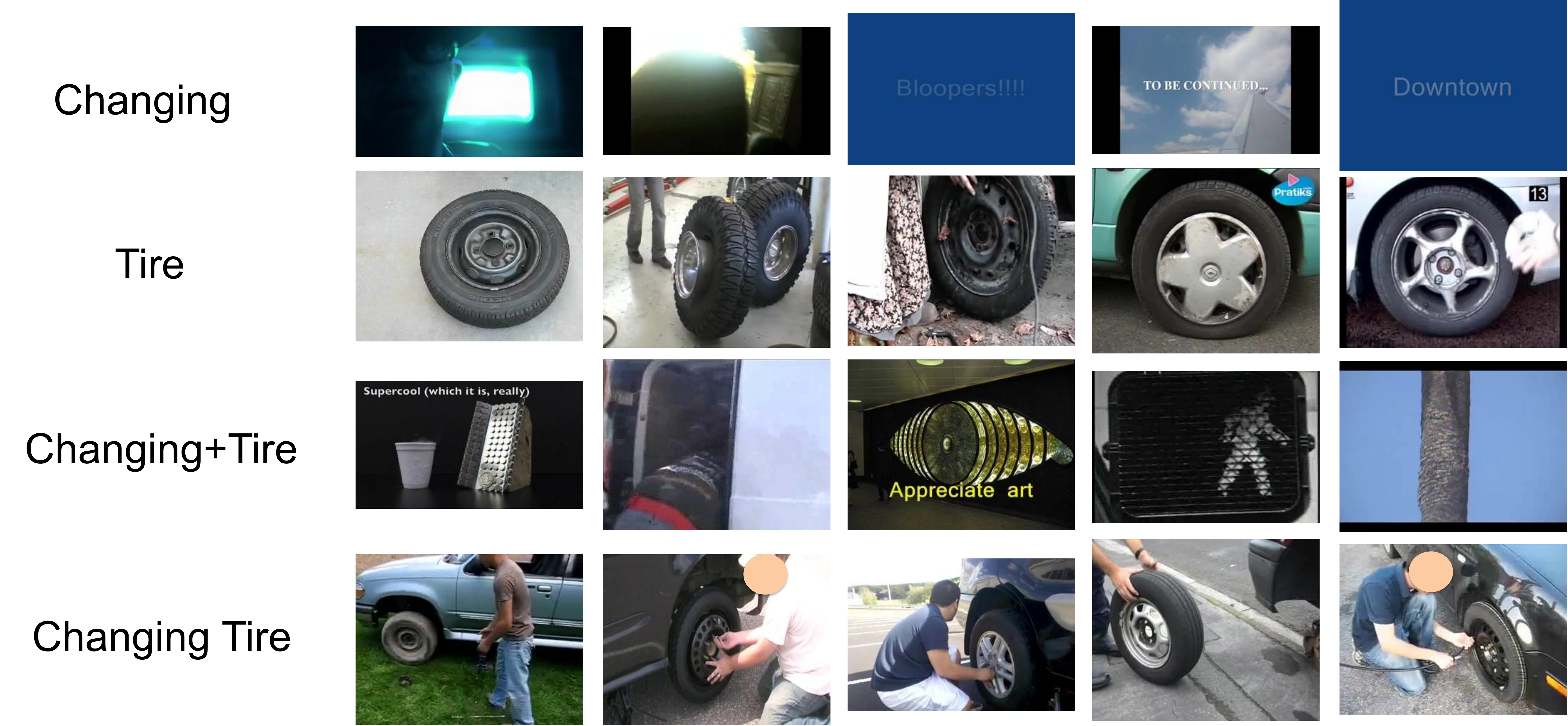}
    \label{fig:stuck+winter}
   \includegraphics[height = 0.142\linewidth]{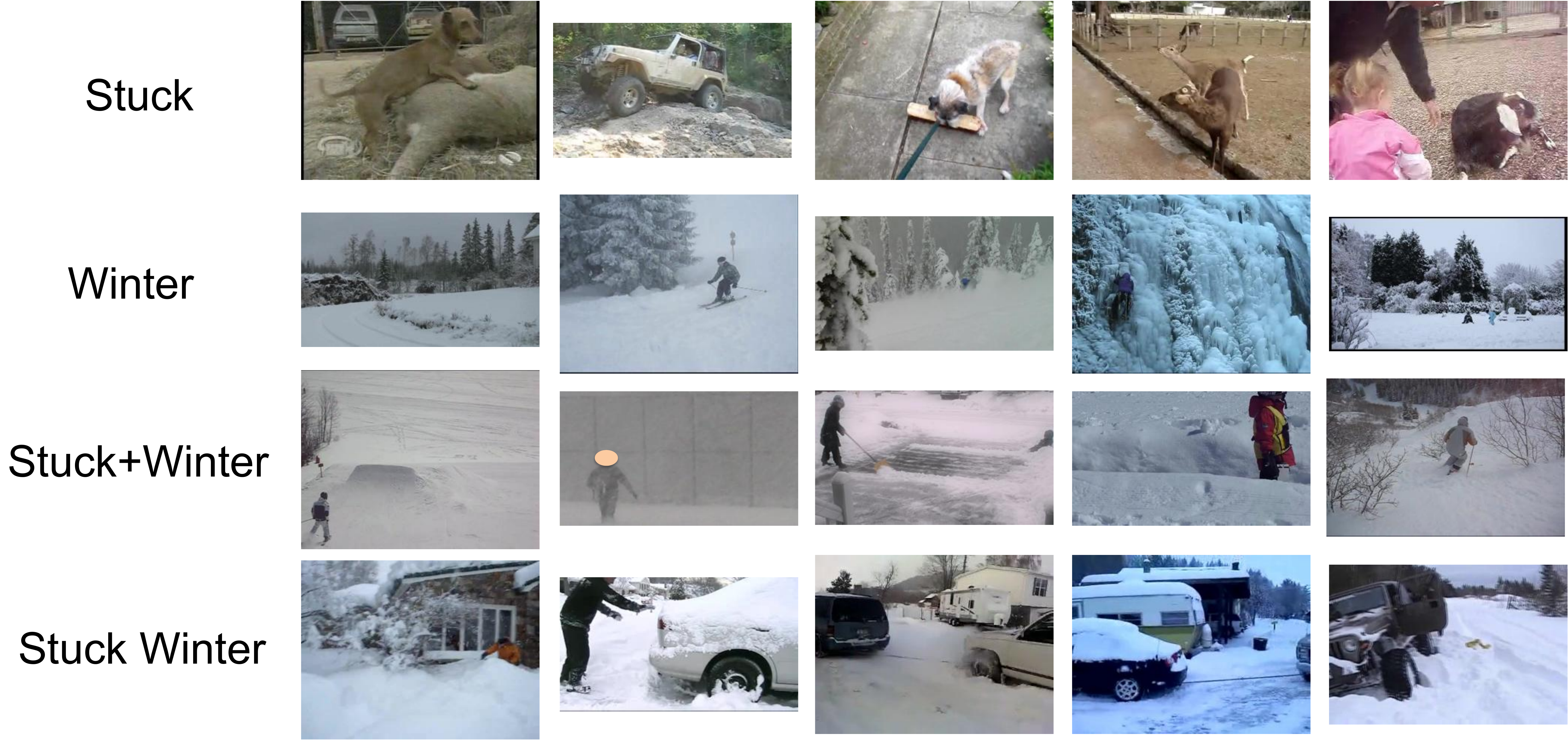}
\end{center}
\caption{Top five ranked videos by different concept detectors trained using web images for three events: (a) {\em attempting a bike trick}, (b) {\em changing a vehicle tire}, (c) {\em getting a vehicle unstuck}. The first and second rows show the results of running unary concepts on test videos. The third row combines two unary concept detectors by adding their scores. The fourth row shows the results of our proposed pair-concept detectors. Pair-concepts are more effective at discovering frames that are more semantically relevant to the event.}
\vspace{-10px}
\label{fig:unary_vs_pair}
\end{figure*}
\section{Overview}
The framework of our algorithm is shown in Fig. \ref{fig:framework}. Given an event defined as a text query, our algorithm retrieves and ranks videos by relevance. The algorithm first constructs a bank of concepts by the approach of \cite{chen2014event} and transforms it into concept pairs. These concept pairs are then pruned by a part of speech model. Each remaining concept pair is used as a text query in a search engine (Google Images), and the returned images are used to train detectors, which are then applied to the test videos. Based on detector responses on test videos, co-occurrence based pruning removes concept pairs that are likely to be outliers. Detectors are then calibrated using a rank based re-scoring method. An instance level pruning method determines how many concept pairs should be observed in a video from the class, discarding the lowest scoring concepts. The scores of the remaining concept pairs are fused to rank the videos. Motion features of the top ranked videos are then used to train a SVM and re-rank the video list. Finally, the top ranked videos are used to re-train the concept detectors, and we use these detectors to re-score the videos.

The following sections describe each part of our approach in detail.

\section{Concept Discovery}

The concept discovery method of \cite{chen2014event} exploits weakly tagged web images and yields an initial list of concepts for an event.  Most of these visual concepts correspond to single words, so they may suffer from ambiguity between linguistic and visual concepts. Consequently, we follow \cite{divvalalearning} by using n-grams to model specific visual meanings of linguistic concepts and \cite{mensinkcosta} by using co-occurrences.  From the top $P$ concepts in the list provided by \cite{chen2014event}, we combine single-word concepts into pairs and retain the phrase concepts to form a new set of concepts. The resulting concepts reduce visual ambiguity and are more informative. We refer to the concepts trained on pairs of words as {\em pair-concepts}.

Fig. \ref{fig:unary_vs_pair} shows the frames ranked highest by the proposed pair-concept detectors, the original concept detectors for single words, and the sum of two independently trained concept detectors on the words constituting the pair-concept. Pair-concept detectors are more relevant to the event than the unary detectors or the sum of two detectors. For example, in Fig. \ref{fig:jump+bicycle}, the event query is {\em attempting a bike trick}, and two related concepts are {\em jump} and {\em bicycle}. The {\em jump} detector can only detect a few instances of jumping, none of which are typical of a bike trick. The {\em bicycle} detector successfully detects bicycles, but most detections are of people riding bicycles instead of performing a bike trick. If the two detectors are combined by adding their scores, some frames with bikes and jump actions are obtained, but they are still not relevant to {\em bike trick}. However,  the {\em jump bicycle} detections are much more relevant to {\em attempting bike trick}--people riding a bicycle are jumping off the ground.

Concepts which do not result in good visual models (e.g., {\em cute water}, {\em dancing blood}) can be identified \cite{divvalalearning,chen2014event}. But, even when concepts lead to good visual models, they might still not be informative (e.g., {\em car truck}, and {\em puppy dog}). Moreover, even if concepts are visual and informative, videos do not always exhibit all concepts related to an event, so expecting all concepts to be observed will reduce retrieval precision. For these reasons, it is not only necessary to select concepts that can be modeled visually, but also to identify subsets of them that are useful to the event retrieval task. We propose three concept pruning schemes to remove bad concepts: pruning based on grammatical parts of speech, pruning based on co-occurrence on test videos, and instance level pruning. The first two schemes remove concepts that are unlikely to be informative, while the last identifies a subset of relevant concepts for each video instance.

\subsection{Part of speech based pruning}
Action centric concepts are effective for video recognition, as shown in \cite{bhattacharyarecognition,sadanand2012action}. Based on this, we require that a pair-concept contain one of three types of action centric words: 1) Nouns that are events, e.g., party, parade; 2) Nouns that are actions, like celebration, trick; 3) Verbs, e.g., dancing, cooking, running. Word types are determined by their
 lexical information and frequency counts provided by WordNet  \cite{miller1995wordnet}. Then, action centric concepts are paired with other concepts that are not action centric to yield the final set of pair-concepts. 

Table \ref{tab:concept} shows the pair-concepts discovered for an event. Qualitatively, these concepts are more semantically relevant to events than the {\em single} word concepts from \cite{chen2014event}. An improvement would be to learn the types of pair-concepts that lead to good event models, based on their parts of speech. However, as our qualitative and quantitative results show, the proposed action-centric pruning rule leads to significant improvements over using all pairs, so we leave data-driven learning for future work.

Pair-concept detectors  are trained automatically using web images. For each concept, 200 images are chosen as positive examples, downloaded by using the concept as the textual query for image search on Google Images. Then, 500 negative examples are randomly chosen from the images of other concepts from all events. Based on the deep features \cite{jia2014caffe} of these examples, the detectors are trained using a RBF kernel SVM using LibSVM \cite{chang2011libsvm} with the default parameters.
\begin{figure*}[]
    \centering
   \includegraphics[width=1\linewidth]{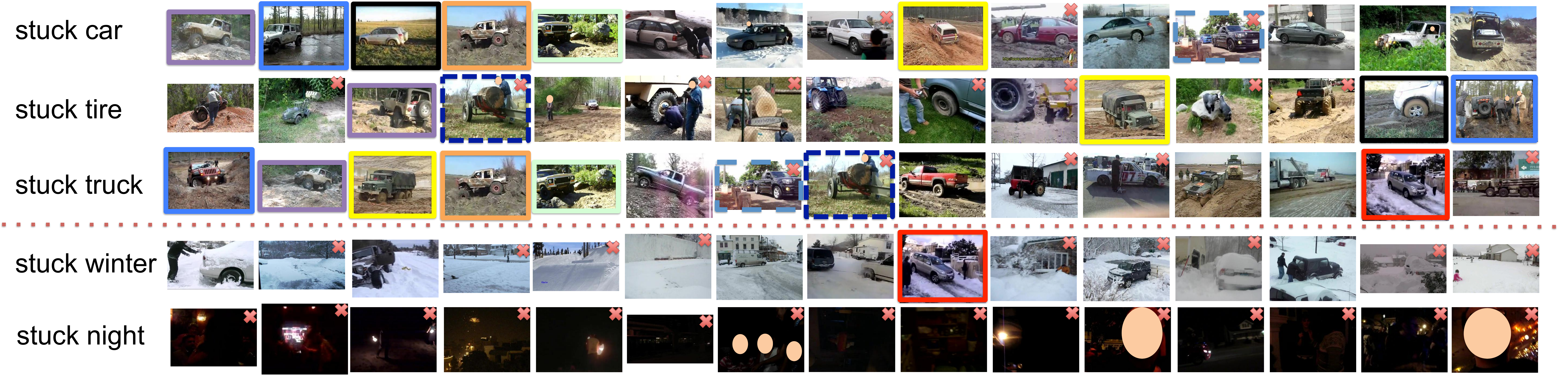}
    \caption{Example of co-occurrence based concept pruning. The five rows correspond to the top 15 videos retrieved by five concept detectors ({\em stuck car}, {\em stuck tire}, {\em stuck truck}, {\em stuck winter}, {\em stuck night}) for detecting the event  {\em getting a vehicle unstuck}. Frames from the same videos are marked with bounding boxes of the same color, and repeating colors across concept detectors denote co-occurrences. For example, the yellow border in rows corresponding to {\em stuck car}, {\em stuck tire}, and {\em stuck truck} signifies that all three concept detectors co-occur in the video represented by the yellow color. The solid boxes denote the positive videos, while the dashed ones are negatives. Stuck night and stuck winter do not co-occur often with other concepts, so they are discarded. Note that the negatives are all marked with red cross in the upper-right corner.}
    \vspace{-10px}

    \label{figure:co-occur}
\end{figure*}

\subsection{Co-occurrence based pruning}

Not all action-centric pair-concepts will be useful, for a number of reasons. First, the process of generating unary-concepts from an event description is uncertain \cite{chen2014event}, and might generate irrelevant ones. Second, even if both unary concepts are relevant individually, they may lead to nonsensical pairs. And finally, even if both unary concepts are relevant, web search sometimes returns irrelevant images which can pollute the training of concept detectors. 

To reduce the influence of visually unrelated and noisy concepts, we search for co-occurrences between detector responses and keep only pair-concepts whose detector outputs co-occur with other pair-concepts at the video level. The intuition is that co-occurrences between good concepts will be more frequent than coincidental co-occurrences between bad concepts. One reason for this is that if two pair-concepts are both relevant to the same complex event, they are more likely to fire in a video of that event. Another reason is that detectors are formed from pairs of concepts, so many pair-concepts will share a unary concept and so are likely to be semantically similar to some extent. For example {\em cleaning kitchen} and {\em washing kitchen} share {\em kitchen}. In other cases, pair-concepts may share visual properties as they are derived for a specific event, for example {\em stuck car} and {\em stuck tire} can co-occur because a tire can be detected along with a car in a frame or in a video.
 
Let $\mathcal{V}$ = $\{V_1, V_2, ... ,V_N\}$ denote the videos in the test dataset, where $V_i$ contains $N_i$ frames $\{V_{i1}, V_{i2}, ..., V_{iN_i}\}$ (sampled from the video for computational reasons). Given $K$ concept detectors $\{D_1,D_2,...,D_K\}$ trained on web-images,  each concept detector is applied to $\mathcal{V}$. For each video $V_i$, an $N_i \times K$ response matrix $\mathbf{S_i}$ is obtained. Each element in $\mathbf{S_i}$ is the confidence of a concept for a frame. After employing a hierarchical temporal pooling strategy (described in section \ref{pooling}) on the response matrix, we obtain a confidence score $s_{ik}$ for the detector $D_k$ applied to video $V_i$. Then for each concept detector $D_k$, we rank the $N$ videos in the test set based on the score $s_{ik}$. Let $L_{k}$ denote the top $M$ ranked videos in the ranking list.We construct a co-occurrence matrix $\mathbf{C}$ as follows:
\begin{equation}
C_{ij} = \begin{cases}
|L_i \cap L_j| / M &  \ \ \ \  1 \leq i,j \leq K, i \neq j \\
0 & \ \ \ \ i = j, \\
\end{cases}
\end{equation}
where $|L_i \cap L_j|$ is the number of videos common to $L_i$ and $L_j$. A concept detector $D_i$ is said to co-occur with another detector $D_j$ if $C_{ij} > t$, where $t$ is between 0 and 1. A concept is discarded if it does not co-occur with $c$ other concepts. 

An example is shown in Fig. \ref{figure:co-occur}. Here, the top 15 ranked videos retrieved by five different concept detectors for the event {\em getting a vehicle unstuck} are shown. The {\em stuck winter} detector co-occurs with other detectors in only one of the top 15 videos, the {\em stuck night} detector does not co-occur with any other detector, so these two detectors are discarded. Also, fewer positive examples of the complex event are retrieved by the two discarded detectors than the other three, suggesting that the co-occurrence based pruning strategy is effective in removing concepts which are outliers.
After pruning some concepts using co-occurrence statistics, we fuse the scores of good concepts by taking the mean score of these concepts and rank the videos in the test set using this score.

\subsection{Instance Level Pruning}
Although many concepts may be relevant to an event, it is not likely that all concepts will occur in a single video. This is because not all complex event instances exhibit all related concepts, and not all concept instances are detected even if they are present (due to computer vision errors). Therefore, computing the mean score of all concept detectors for ranking is not a good solution. So, we need to predict an event when only a subset of these concepts is observed. However, the subset is video instance specific and knowing all possible subsets a priori is not feasible with no training samples. Even though these subsets cannot be determined, we can estimate the average cardinality of the set based on the detector responses observed for the top $M$ ranked videos after computing the mean score of detectors. For each event, the number of relevant concepts is estimated as:
\begin{equation}
\mathit{N_{r}} = K - min(\lceil \frac{\sum_{k = 1}^{K}\sum_{i = 1}^{M}\mathbf{1}(s_{ik} < T)}{M} \rceil, \lambda)
\label{eq:prune}
\end{equation}
where $\mathbf{1}(\cdot)$ is the indicator function--it will be 1 if the confidence score of concept $k$ present in video $V_i$ is less than a detection threshold $T$ (i.e., detector $D_k$ does not detect the concept $k$ in the video $V_i$) and 0 otherwise. $\lceil \cdot \rceil$ is the ceiling function, and $\lambda$ is a regularizer to control the maximum number of concepts to be pruned for an event. This equation computes the average number of detected concepts in the top ranked videos. When combining the concept scores, we keep only the top $N_{r}$ responses and discard the rest.
\begin{table*}[ht]
\small
\centering
\caption{Concepts discovered after different pruning strategies}
\begin{tabular}{|p{50pt} | p{75pt} | p{300pt} |}
\hline
\textbf{Event} & \multicolumn{2}{c|}{\textbf{Discovered Concepts}} \\
\hline
\multirow{3}{50pt}{\\\textit{Working on a metal crafts project}} & Initial Concepts & art, bridge, iron, metal, new york, new york city, united state, work, worker \\  \cline{2-3}
& After part of speech based pruning & iron art, iron bridge, iron craft, metal art, metal bridge, metal craft, new york, new york city, united state, work iron, work metal, work worker, worker art, worker bridge, worker craft \\  \cline{2-3}
& After co-occurrence based pruning & iron art, iron craft, metal art, metal bridge, metal craft, work iron, work metal\\ \hline
\multirow{3}{50pt}{\\\textit{Dog show}} & Initial Concepts & animal, breed, car, cat, dog, dog show, flower, pet, puppy, show \\  \cline{2-3}
& After part of speech based pruning & animal pet, breed animal, breed car, breed cat, breed dog, breed flower, breed puppy, car pet, cat pet, dog pet, dog show, flower pet, puppy pet, show animal, show car, show cat, show dog, show flower, show puppy \\  \cline{2-3}
& After co-occurrence based pruning & animal pet, breed animal, breed car, breed cat, breed dog, breed puppy, cat pet, dog pet, dog show, puppy pet, show cat, show dog, show puppy \\ \hline
\multirow{3}{50pt}{\\\textit{Parade}} & Initial Concepts & city, gay, gay pride, gay pride parade, new york, new york city, nyc event, parade, people, pride \\  \cline{2-3}
& After part of speech based pruning & city parade, gay city, gay people, gay pride, gay pride parade, new york, new york city, nyc event, people parade, pride parade \\  \cline{2-3}
& After co-occurrence based pruning & city parade, gay pride, gay pride parade, people parade, pride parade\\ \hline
\end{tabular}
\label{tab:concept}
\end{table*}
\section{Hierarchical Temporal Pooling}
\label{pooling}
Our concept detectors are frame-based, so we need a strategy to model the temporal properties of videos. A common strategy is to treat the video as a bag, pooling all responses by the average or max operator. However max-pooling tends to amplify false positives; on the other extreme, average pooling would be robust against spurious detections, but expecting a concept to be detected in many frames of a video is not realistic and would lead to false negatives. As a compromise, we propose hierarchical temporal pooling, where we perform max pooling within sub-clips and average over sub-clips over a range of scales. Note that the top level of this hierarchy corresponds to max pooling, the bottom level corresponds to average pooling, and the remaining levels correspond to something in-between. The score for a concept $k$ in video $V_i$ is computed as follows,
\begin{equation}
s_{ik} = \sum_{n = 1}^{l} \sum_{j = 1}^{n} \frac{m_{nj}}{n}
\label{eq:pool}
\end{equation}
where, $l$ is the maximum number of parts into which a video is partitioned (a scale at which the video is analyzed), $m_{nj}$ is the max pooling score of the detector in part $j$ of the video partitioned into $n$ equal parts. Temporal pooling has been widely used in action recognition \cite{laptev2008learning} for representing space-time features. In contrast, we perform temporal pooling over SVM scores, instead of pooling low level features.

\section{Domain Adaptation}
\textbf{Score Calibration.} 
The detectors are trained on web-images, so their scores are not reliable because of the domain shift between the web and video domains. In addition, each detector may have different response characteristics on videos, e.g., one detector is generic and has a high response for many videos, while another detector is specific and has a high response only for a few videos. Thus we calibrate their responses before fusion as follows :
\begin{equation}
s'_{ik} = \frac{1}{1 + exp(\frac{\mathit{R_k}(s_{ik}) - u}{u})}
\label{eq:calib}
\end{equation}
where, $s'$ is the calibrated score, $\mathit{R_k}$ is the rank of video $V_i$ when generating the rank list only using concept detector $D_k$, and $u$ controls the decay factor in the exponential. This re-scoring function not only calibrates raw detector scores, but it also gives much higher score to highly ranked samples while ignoring the lower ranked ones, which is appropriate for retrieval.

\textbf{Detector Retraining.}
Based on the domain adaptation approach of \cite{jiang2014zero}, we use pseudo-relevance from top-ranked videos to improve performance. Since web-detectors only capture static scene/object cues in a video, it is beneficial to extract Fisher Vectors (FV) on Improved Dense Trajectory (IDT) features \cite{wang2013action} to capture the motion cues. Based on the rank list obtained from concept detectors, we train a linear SVM using LIBLINEAR \cite{fan2008liblinear} on the top ranked videos using the extracted Fisher Vectors. The lowest ranked videos are used as negative samples. These detectors are applied again on the test videos. Finally, we use late fusion to combine the detection scores obtained using motion features with web-detectors.

We further adapt the concept detectors to the video domain by retraining them on frames from top-ranked videos. For each detector, we obtain frames with the highest response in the top ranked videos (after fusion with motion features) to train a concept detector (with the constraint that similar frames should not be selected twice to encourage diversity). We then repeat the process of co-occurrence based pruning, instance level pruning and rank based calibration to fuse the scores for the new concept detectors. Finally, the video scores are updated by summing the fused scores (original concept detectors + IDT) and the scores of adapted concept detectors.\begin{figure*}[htbp]
\begin{center}
\includegraphics[width=1\textwidth]{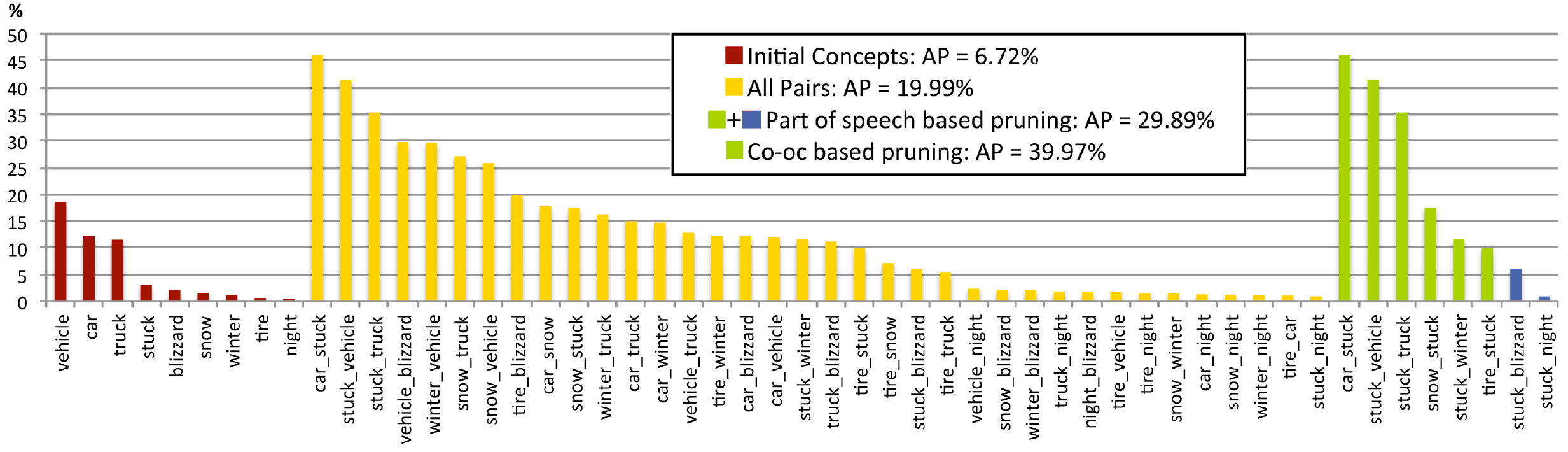}
\caption{Average Precision (AP) scores of initial concepts, all pair-concepts, the concepts after part of speech based and co-occurrence pruning are shown for the event "Getting a vehicle unstuck". AP after combining the concepts is also reported. Note that part of speech pruning helps in removing many pair-concepts with low AP. Moreover, co-occurrence based pruning removes the two lowest performing pair-concepts and improves the AP after part of speech pruning significantly. }
\vspace{-20px}
\label{fig:indscore}
\end{center}
\end{figure*}
\section{Experiments and Results}
We perform experiments on the challenging TRECVID Multimedia Event Detection (MED) 2013 dataset. We first verify the effectiveness of each component of our approach, and then show the improvement on the EK0 dataset by comparing with state-of-the-art methods.
\subsection{Dataset and Implementation Details}
The TRECVID MED 2013 EK0 dataset consists of unconstrained Internet videos collected by the Linguistic Data Consortium from various Internet video hosting sites. Each video contains only one complex event or content not related to any event. There are 20 complex events in total in this dataset, with ids 6-15 and 21-30. These event videos together with background videos (around 23,000 videos), form a test set of 24,957 videos. In the EK0 setting, no ground-truth positives training videos are available. We apply our algorithm on the test videos, and mAP score is calculated based on the video ranking.
\begin{figure*}[htbp]
\begin{center}
\includegraphics[width=1\textwidth]{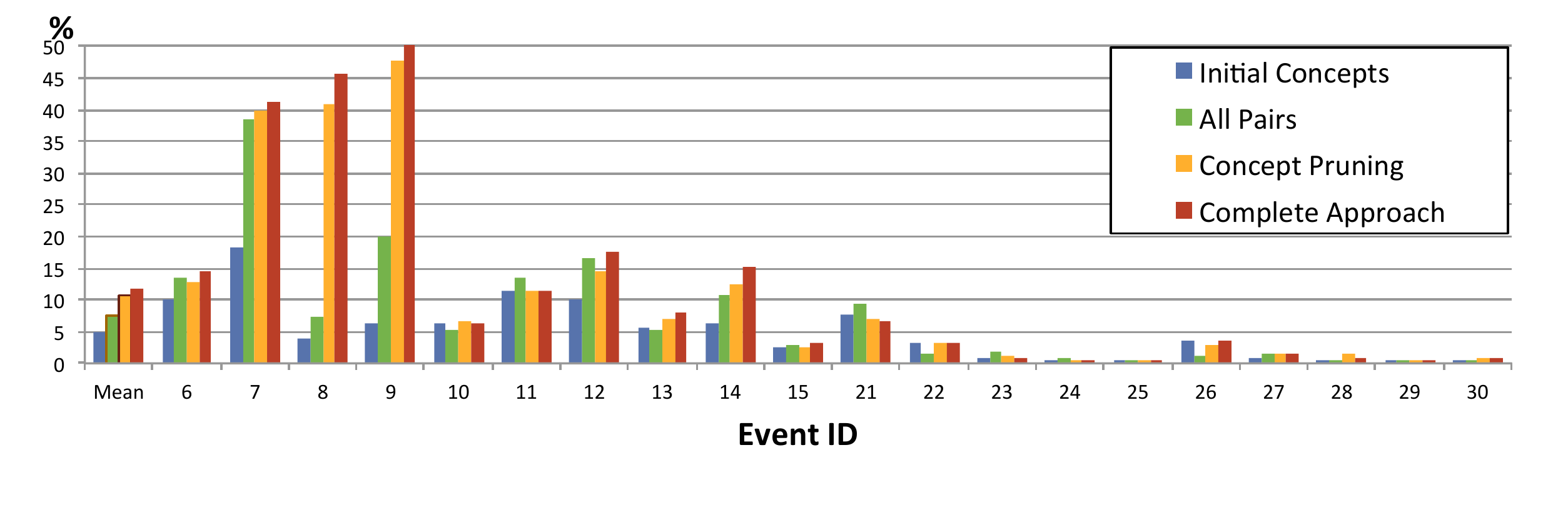}
\vspace{-30px}
\caption{Mean average precision (mAP) scores for the events on the MED13 EK0 dataset. By pruning concepts that are not useful for retrieving the complex event of interest, our approach progressively improves the utility of the remaining.}
\vspace{-10px}
\label{fig:bar}
\end{center}
\end{figure*}

For each event in EK0 dataset, we choose the top 10 concepts (i.e., $P = 10$) in the list provided by \cite{chen2014event} and transform them into pair-concepts. The web image data on which concept detectors are trained is obtained by image search on Images using each pair-concept as a query. The \emph{Type} option is set to \emph{Photo} to filter out irrelevant cartoon images. We downloaded around 200 images for each concept pair query. We sample each video every two seconds to obtain a set of frames. Then, we use Caffe \cite{jia2014caffe}  to extract deep features on all the frames and web images, by using the model pre-trained on ImageNet. We used the fc7 layer after dropout, which generates a 4,096 dimensional feature for each video frame or image. The hyper-parameters to determine if a concept co-occurs with another $t$, length of the intersection list $M$, regularization constant $\lambda$, detector threshold $N_r$ are selected based on leave one-event out cross validation, since they should be validated with event categories different from the one being retrieved. We found hyper-parameters to be robust after doing sensitivity analysis. The number of levels $l$ in hierarchical temporal pooling was set to 5. The Fisher Vectors of the top 50 ranked videos and the bottom 5,000 ranked videos are used to train a linear SVM.

\begin{table}[ht]
\centering
\caption{Comparative results}
\begin{tabular}{|c|c|c|}
\hline
\textbf{Pre-Defined} & \textbf{Method} & \textbf{mAP} \\
\hline
No & Concept Discovery \cite{chen2014event} & 2.3\% \\ 
Yes & SIN/DCNN \cite{jiang2014zero} & 2.5\% \\
Yes & CD+WSC \cite{wu2014zero} & 6.12\% \\
Yes & Composite Concept \cite{habibian2014composite}& 6.39\% \\ \hline
 & {\em Initial} concepts & 4.91\% \\
 & {\em All} Pair-concepts & 7.54\% \\
No & +Part of speech pruning & 8.61\% \\
 & +Cooc \& inst pruning & 10.85\% \\
& \textbf{+Adaptation} & \textbf{11.81\%} \\ \hline
\end{tabular}
\label{tab:comparison}
\end{table}

\subsection{Evaluation on MED 13 EK0}
Table \ref{tab:concept} shows the initial list of concepts, the concepts that remain after part of speech based pruning, and the concepts that remain after co-occurrence based pruning for three different events. Although the initial concepts are related to the event, web queries corresponding to them would provide very generic search results. Since we have 10 unary concepts per event, there are 45 unique pair-concept detectors for each event. Approximately 10-20 pair-concepts remain after part of speech based pruning. This helps to reduce the computational burden significantly and also prunes away noisy pairs. Finally, co-occurrence based pruning discards additional outliers in the remaining pair-concepts. 

Table \ref{tab:comparison} shows the results of our method on the TRECVID EK0 dataset. We observe significant performance gains (5.4\% - 11.81\% vs 6.39\%) over other vision based methods which do not use any training samples. Our performance is almost 2-5 times their mAP. Note that the methods based on pre-defined concepts must bridge the semantic gap between the query specification and the pre-defined concept set. On the other hand, we leverage the web to discover concepts. Our approach follows the same protocol as \cite{chen2014event} which performs the same task. Using the same initial concepts as \cite{chen2014event}, our method obtains 5 times the mAP as that of \cite{chen2014event}. Fig. \ref{fig:bar} shows the effect of each stage in the pipeline. Replacing the initial set of concepts by action based pair-concepts provides the maximum gain in performance of $\sim$3.7\% (4.9\% to 8.61\%). Next, co-occurrence based pruning improves the mAP by 1.8\% (8.61\% to 10.4\%). Calibration of detectors and instance level pruning improves the mAP score to 10.85\%. Finally, adapting each detector on the test dataset and using motion information allows us reach a mAP of 11.81\%. The performance is low for events 21 to 30 because there are only $\sim$25 videos for these events while events 6-15 have around 150 videos each in the test set.

To illustrate that the proposed pruning methods remove concepts with low AP, in Fig \ref{fig:indscore} we plot AP scores of initial unary concepts, all pair-concepts, part of speech based concepts and the concepts after co-occurrence based pruning. Note that almost 50\% of pair-concepts had an average precision below 10\% before pruning. After part of speech and co-occurrence based pruning, our approach is able to remove all these low scoring concepts in this example.

We would note that Hierarchical Temporal Pooling provides significant improvement in performance for this task. In Table \ref{tab:pool}, we show mAP scores for different pooling methods for initial, pair-concepts and after concept pruning (before Detector Retraining). It is clear that Hierarchical Temporal Pooling improves performance in all three cases. We also observe that concepts after pruning have best performance across all pooling methods .

\begin{table}[ht]
\centering
\caption{Pooling Results}
\begin{tabular}{|c|c|c|c|}
\hline
 & \textbf{Initial} & \textbf{All Pairs} & \textbf{After Pruning}\\
\hline
Avg. Pooling & 2.84\% & 4.54\% & 5.94\% \\ 
Max. Pooling & 4.45\% & 6.87\% & 9.01\% \\
Hierarchical & 4.91\% & 7.54\% & 10.85\% \\
\hline
\end{tabular}
\label{tab:pool}
\end{table}
\vspace{-5px}

\section{Conclusion}
We demonstrated that carefully pruning concepts can significantly improve performance for event retrieval when no training instances of an event are available, because even if concepts are visually salient, they may not be relevant to a specific event or video. Our approach does not require manual annotation, as it obtains weakly annotated data through web search, and is able to automatically calibrate and adapt trained concepts to new domains.

\section*{Acknowledgement}
This work is supported by the Intelligence Advanced Research Projects Activity (IARPA) via the Department of Interior National Business Center contract number D11PC20071. The U.S. Government is authorized to reproduce and distribute reprints for Governmental purposes not with standing any copyright annotation thereon. The views and conclusions contained herein are those of the authors and should not be interpreted as necessarily representing the official policies or endorsements, either expressed or implied, of IARPA, DoI/NBC, or the U.S. Government. The authors would like to thank Yin Cui for providing the initial concepts. The authors acknowledge the University of Maryland supercomputing resources \url{http://www.it.umd.edu/hpcc} made available for conducting the research reported in this paper.
{\small
\bibliographystyle{ieee}
\bibliography{egbib}
}

\end{document}